\documentclass[sigconf,nonacm,screen,pbalance]{acmart}
  
  \AtBeginDocument{%
    }

  \usepackage{algorithm}
  \usepackage{algpseudocodex}
  \usepackage{graphicx}
  \usepackage{textcomp}
  \usepackage{xcolor}
  \usepackage{float}
  \usepackage{subcaption}
  \usepackage{booktabs}
  \usepackage{amsmath}
  \usepackage{multirow}
  \usepackage{colortbl}
  \definecolor{gold}{HTML}{FFD700}
  \definecolor{silver}{HTML}{C0C0C0}
  \definecolor{bronze}{HTML}{CD7F32}
  \usepackage{tikz}
  \usetikzlibrary{positioning}
  
  \usepackage{booktabs}
  \usepackage{enumitem}
  \usepackage{hyperref}
  \usepackage[para]{footmisc}
  \usepackage{pifont}
  \usepackage{listings}
  \algnewcommand{\IfReturn}[2]{%
    \If{#1} \textbf{return} #2; \EndIf%
  }
  \setcopyright{acmlicensed}
  \copyrightyear{2025}
  \acmYear{2025}
  \acmDOI{XXXXXXX.XXXXXXX}
  \acmISBN{XXX-X-XXXX-XXXX-X/2026/07}

\begin{document}
  \title[STG]{Structured Testbench Generation for LLM-Driven HDL Design and Verification-Oriented Data Curation}

  \author{
    En-Ming~Huang$^{1}$, Yu-Hung~Kao$^{1}$, Ren-Hao~Deng$^{1}$, Wei-Po~Hsin$^{1}$,
    Yao-Ting~Hsieh$^{2}$, Cheng~Liang$^{1}$, Hsiang-Yu~Tsou$^{1}$, Mu-Chi~Chen$^{1}$,
    Yu-Kai~Hung$^{1}$, Shao-Chun~Ho$^{1}$, Po-Hsuang Huang$^{1}$, Shih-Hao~Hung$^{1}$,
    H.T.~Kung$^{3}$
  }
  \affiliation{%
    \institution{$^{1}$National Taiwan University, $^{2}$Academia Sinica, $^{3}$Harvard University}
    \city{}
    \country{}
  }
  \email{{r13922078@csie.ntu.edu.tw, hungsh@csie.ntu.edu.tw, kung@harvard.edu}}

  \renewcommand{\shortauthors}{Huang et al.}

  \authorsaddresses{Contact authors: \email{r13922078@csie.ntu.edu.tw}, \email{hungsh@csie.ntu.edu.tw}}
  \begin{abstract}
Automated testbench generation has become a critical bottleneck in large language model (LLM)-driven Register Transfer Level (RTL) workflows, where large numbers of candidate designs must be verified rapidly and reliably. Existing prompt-based approaches treat testbench generation as unconstrained code synthesis, yielding stochastic outputs with high token cost, low reproducibility, and insufficient coverage. To address this gap, we present STG, a Structured Testbench Generation framework that exploits the inherent structure of hardware designs to generate deterministic testbenches. As a direct verification tool, STG runs $720\times$ faster than an iterative LLM-based testbench generation flow and higher rate of successful compilation, achieves higher coverage, and reduces false-pass verdicts on incorrect DUTs. STG also helps identify errors in RTL generation benchmarks by exposing faulty benchmark testbenches. As a data curation engine, it is $11\times$ faster than LLM-based filtering on a single CPU core with $127\times$ less energy, and the resulting distilled models provide state-of-the-art performance in our multi-benchmark evaluation. As a test-time scaling oracle, it reduces node count by 14-47\%. Our models are available at \url{https://huggingface.co/collections/AS-SiliconMind/siliconmind-v12}.
\end{abstract}


  \maketitle

  \section{Introduction}
\label{sec:intro}

Functional verification remains one of the most labor-intensive stages of hardware design. As Register Transfer Level (RTL) designs grow in complexity, constructing testbenches that expose corner cases requires substantial manual effort. Prior work has explored automated stimulus generation through finite-state machine (FSM) modeling~\cite{Chow1978FSM}, coverage-guided simulation~\cite{Amla2001BiasedRandom}, and probabilistic methods~\cite{Ferens2003Bayesian}, yet practical testbench development remains a major bottleneck.
This challenge becomes more acute in the era of large language models (LLMs), where hardware description language (HDL) code can now be generated at scale from natural languages. Recent LLM-driven hardware design systems use generated verification artifacts for spec-to-RTL evaluation, dataset construction, and test-time feedback~\cite{Liu2024AutoBench,Liu2025CorrectBench,Liu2025ConfiBench,Yao2025CodeV,Chen2026SiliconMindV1}. In such settings, verification is no longer only a downstream design step; it becomes a core mechanism for validating generated HDL artifacts, filtering low-quality outputs, and organizing data for subsequent model improvement.

Nevertheless, we observe that existing LLM-based testbench generation methods~\cite{Liu2024AutoBench,Liu2025CorrectBench,Liu2025ConfiBench,teng2025verirl} are framed as unconstrained code generation, which leads to two limitations. First, as testbenches are generated through a stochastic process by LLMs, improving reliability requires iterative prompting or ensemble generation, thereby increasing token cost. Second, this formulation overlooks the structured nature of simulation-based verification: module instantiation, output checking, and reporting can be generated directly, while the core challenge reduces to producing high-coverage stimuli.

These limitations are amplified by several emerging demands in LLM-driven HDL workflows. Test-time scaling techniques---such as Monte Carlo Tree Search (MCTS)-based workflow search~\cite{wei2026vflow} and evolutionary refinement~\cite{novikov2025alphaevolve,min2026revolution}---now place verification inside an iterative optimization loop, where each candidate revision must be evaluated quickly and reliably before the search can proceed; a noisy verification signal directly degrades search efficiency and quality. Model-distillation pipelines generate large numbers of candidate DUTs that must be validated before they can serve as training data~\cite{QiMeng2025CodeVR1,teng2025verirl,Chen2026SiliconMindV1}; weak or unstable testbenches misclassify candidate designs, introduce noisy labels, and reduce the value of the curated dataset. This cost pressure intensifies further as LLM training moves toward continuous learning, where models are iteratively retrained on freshly data~\cite{2025continuelearningsurvey}, and as distilled smaller models find new roles such as speculative-decoding draft models~\cite{2023specdec} that accelerate large-model inference. All settings demand a low-cost verification mechanism that scales to large numbers of candidates.

In this work, we present STG, a Structured Testbench Generation framework that combines lightweight HDL analysis with template-based rendering to produce testbenches deterministically for both combinational and general sequential designs. STG is designed to serve as a general-purpose verification backend for LLM-driven HDL workflows, supporting three closely related scenarios: (i)~direct RTL verification, in which candidates are verified against a golden reference; (ii)~verification-oriented dataset curation, as large batches of generated artifacts must be filtered before use as training data; and (iii)~test-time scaling, where reliable verification feedback must be provided at every iteration of an LLM-guided refinement loop.

We evaluate STG and its applications on Verilog generation benchmarks~\cite{Liu2023VerilogEval,Thakur2024RevisitingVerilogEval}. STG generates testbenches $720\times$ faster than an iterative LLM-based testbench generation pipeline~\cite{Liu2025ConfiBench}, with higher line and toggle coverage and fewer false-pass verdicts on incorrect DUTs. For data curation, STG is $10.6\times$ faster on a CPU core with $127\times$ less energy than LLM-based filtering, and our simple supervised fine-tuning (SFT) pipeline yields competitive or superior results in our multi-benchmark evaluation~\cite{Liu2023VerilogEval,Thakur2024RevisitingVerilogEval,pinckney2025cvdp,lu2024rtllm} while using less training data than recent specialized baselines~\cite{teng2025verirl,QiMeng2025CodeVR1,Chen2026SiliconMindV1}. In test-time scaling, STG reduces solved-problem node count by 14--47\% on existing LLMs~\cite{Chen2026SiliconMindV1,openai2025gptoss,Guo2025deepseek}. We also identify and correct a systematic race condition in VerilogEval's testbenches~\cite{Liu2023VerilogEval,Thakur2024RevisitingVerilogEval} through STG's deterministic generation and human inspection. Our results further indicate that the effectiveness of recent complex training and reinforcement learning workflows~\cite{teng2025verirl,QiMeng2025CodeVR1} remains questionable.

Our main contributions are threefold:
\textbf{(1)} We present STG, a deterministic and structure-aware testbench generation
framework for RTL verification that improves over prompt-based LLM testbench
generation in efficiency, coverage, and reliability.
\textbf{(2)} We show that STG enables efficient verification-oriented data curation
and supports strong distilled RTL generation models using a simple pipeline.
\textbf{(3)} We demonstrate that STG serves as an effective verification backend for
LLM-driven RTL refinement, improving search quality and efficiency across multiple
backbone models.
Out models are available at \url{https://huggingface.co/collections/AS-SiliconMind/siliconmind-v12}.

The remainder of this paper is organized as follows. Section~\ref{sec:background} reviews related work on LLM-based testbench generation and verification-oriented RTL workflows. Section~\ref{sec:methodology} introduces the STG framework. Section~\ref{sec:applications} describes the applications of STG. Section~\ref{sec:experiments} presents experimental results, and Section~\ref{sec:conc} concludes.

  \section{Problem Definition and Background}
\label{sec:background}

This section formalizes the verification problem addressed by STG. We define the known-reference RTL verification setting and its requirements, review existing LLM-based testbench-generation workflows, and discuss how test-time scaling and data curation create additional demands on testbench quality.

\subsection{Known-Reference RTL Verification}
\label{subsec:known-ref}

We consider the known-reference RTL verification setting. Given a design under test (DUT) $D$ and a trusted golden implementation $G$, the objective is to automatically construct a testbench $T$ that applies effective stimuli to $D$, compares its behavior against $G$, and determines whether $D$ is functionally correct. The generated testbench must satisfy three practical requirements: (i)~produce trustworthy pass/fail judgments, (ii)~achieve high behavioral coverage, particularly for control- and state-dependent behaviors, and (iii)~incur low generation cost so that it scales to large batches of RTL candidates.

This setting is broadly applicable in current LLM-driven RTL workflows, where golden references are routinely available: RTL generation benchmarks ship with reference implementations~\cite{Liu2023VerilogEval,Thakur2024RevisitingVerilogEval,lu2024rtllm,pinckney2025cvdp}, test-time scaling systems generate candidates against a known specification~\cite{wei2026vflow,min2026revolution}, and data-curation pipelines filter LLM outputs against trusted solutions~\cite{Yao2025CodeV,QiMeng2025CodeVR1,Chen2026SiliconMindV1}. The known-reference assumption therefore covers the three use cases introduced in Section~\ref{sec:intro}: \emph{direct RTL verification}, \emph{verification-oriented data curation}, and \emph{test-time scaling}.
We therefore formulate the target problem as ``structured testbench generation for verification-oriented classification'': given $D$ and $G$, generate a testbench that reveals meaningful behaviors of $D$, produces a trustworthy pass/fail decision, and scales to large numbers of generated RTL candidates.

\subsection{LLM-Based Testbench-Generation Methods}
\label{subsec:llm-tb}

A line of prior work---AutoBench~\cite{Liu2024AutoBench}, CorrectBench~\cite{Liu2025CorrectBench}, and ConfiBench~\cite{Liu2025ConfiBench}---tackles the open-ended setting where no trusted reference exists, and the LLM must synthesize both stimulus and a \emph{silver reference} oracle from scratch. A \emph{silver reference} is an alternative implementation of the same specification produced by an LLM (e.g., a behavioral model in C++ or Python), used as a substitute oracle when no authoritative golden reference is available. While these methods progressively improve generation quality through self-correction and ensembling, they share a fundamental ambiguity: when the DUT and the oracle are both produced by the same stochastic process, a mismatch cannot be unambiguously attributed to a bug in the DUT versus an error in the oracle, making the pass/fail verdict inherently unreliable. The known-reference setting assumed in this work eliminates this ambiguity by assuming a trusted golden reference $G$ in hand, so any discrepancy is definitively a DUT fault. This shifts the problem from open-ended code synthesis to efficient, high-coverage stimulus generation.

\subsection{Test-Time Scaling and Verification-Oriented Data Curation}
\label{subsec:data-curation}

The need for efficient known-reference verification is amplified by two recent trends in LLM-driven RTL generation that both rely heavily on verification quality: test-time scaling and verification-oriented data curation.

\textbf{\textit{Test-time scaling.}}
Recent LLM-based RTL generation systems have moved beyond one-shot prompting toward iterative search and refinement at inference time, placing verification inside the optimization loop rather than after it~\cite{wei2026vflow, min2026revolution, Dong2025ScaleRTL}. The architectural patterns vary but all share a common requirement: at every iteration, the system must evaluate each candidate and use the result to decide what to generate next. This turns the testbench into a performance-critical component of the generation process itself. A noisy or unreliable verification signal can cause the search to retain faulty candidates, reject correct ones, or waste iterations on ambiguous feedback. The testbench must therefore be not only correct but also fast to generate, deterministic, and informative enough to distinguish partially correct designs from wholly incorrect ones.

\textbf{\textit{Verification-oriented data curation.}}
A parallel development is the growing use of model distillation and reinforcement learning to train small, thinking models that are specialized for RTL generation~\cite{teng2025verirl,Yao2025CodeV,QiMeng2025CodeVR1,Chen2026SiliconMindV1}. These pipelines produce large numbers of candidate DUTs, often paired with reasoning traces or auxiliary artifacts, that must be validated and classified before they can serve as training data. The verification artifacts in the filtering stage, however, are still commonly handled through prompt-based LLMs, which are expensive to scale when screening large datasets. Weak or unstable testbenches at this stage can also misclassify candidate designs, introduce noisy labels, and degrade the quality of the curated dataset~\cite{QiMeng2025CodeVR1,Chen2026SiliconMindV1}. Currently, no mechanism exists that can filter large numbers of candidates cheaply and reproducibly without requiring per-task LLM invocation~\cite{QiMeng2025CodeVR1,teng2025verirl,Chen2026SiliconMindV1}.

Both trends redefine the role of verification in LLM-driven RTL pipelines. Verification no longer serves solely to judge whether a generated DUT is correct; it also provides the feedback signal inside search loops and the quality gate for training-data construction. The verification engine thus becomes part of the core infrastructure for model improvement, making low-cost, reproducible, and behaviorally meaningful testbench generation especially valuable.
  \section{STG: Structured Testbench Generation}
\label{sec:methodology}
\begin{figure*}[t]
    \centering
    \includegraphics[width=0.85\linewidth]{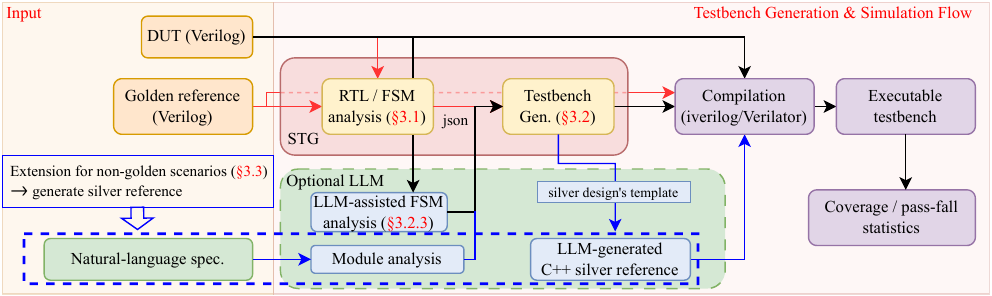}
    \vspace{-1.3em}
    \caption{Overall workflow of STG. STG is mainly designed for the condition which both DUT and golden reference are available, while still can be extended to the silver-reference setting discussed in $\S$\ref{subsec:llm-tb}. The {\color{red}red lines} correspond to the main workflow of STG; the {\color{blue}blue lines} indicate the extension to the silver-reference setting. The black lines are common steps for both settings.}
    \label{fig:stg-flow}
    \vspace{-1em}
\end{figure*}

Fig.~\ref{fig:stg-flow} shows the overall flow of STG. The framework takes both the DUT and golden reference as input. STG first analyzes the HDL structure, then generates a testbench deterministically according to the detected design type, and finally compiles and executes the testbench to obtain pass/fail statistics and coverage information. The testbench is rendered from parameterized \textit{Jinja} templates populated with the extracted module information, including port lists, signal roles, and design-type-specific parameters.

In this section, we first describe the module parsing and analysis process, which extracts the necessary information from the HDL code to guide the generation. We then present the different stimulus-generation strategies for combinational, general sequential, and FSM-dominated designs. Finally, we discuss how STG can be extended to the setting where no trusted golden reference is available and an LLM-generated silver reference is used instead, as in the workflows discussed in Section~\ref{subsec:llm-tb}.

\subsection{Module Parsing and Analysis}
\label{subsec:module-parsing}

STG operates in two modes. In \emph{automatic mode}, the framework analyzes the DUT entirely through heuristics and lightweight LLM queries, requiring no human intervention. This mode is designed for large-scale data curation where thousands of modules must be processed without manual effort. In \emph{interactive mode}, a user may supply additional hints---such as explicit signal roles or design-type overrides---to improve accuracy for a specific verification task.

\textbf{\textit{Top-module identification.}}
STG parses all module instantiations in the input files using Icarus Verilog~\cite{williams2002icarus} to construct a module-instantiation directed acyclic graph. The top module is identified as the root node of this graph. When multiple roots exist (e.g., if utility modules are also provided), STG selects the root with the most descendant nodes as the top module.

\textbf{\textit{Design-type classification.}}
\begin{table}[t]
  \centering
  \caption{Design-type classification and corresponding testbench generation strategy.}
  \label{tab:design-type}
  \vspace{-1em}
  \footnotesize
  \begin{tabular*}{\columnwidth}{@{\extracolsep{\fill}}lll@{}}
    \toprule
    \textbf{Design type} & \textbf{Detection criterion} & \textbf{Strategy} \\
    \midrule
    Combinational      & No clock in port list        & Exhaustive / random (\S\ref{subsec:comb}) \\
    General sequential & Clock present, no FSM        & Phased random (\S\ref{subsec:seq}) \\
    FSM-dominated      & Clock present, FSM detected  & Transition-guided (\S\ref{subsec:fsm}) \\
    \bottomrule
  \end{tabular*}
  \vspace{-2em}
\end{table}

STG classifies each design into one of three categories, \emph{combinational}, \emph{general sequential}, or \emph{FSM-dominated}, since each category requires a different strategy for stimulus generation (detailed in Sections~\ref{subsec:comb}--\ref{subsec:fsm}). The classification proceeds as follows. First, STG checks whether any clock signal is present in the port list. If no clock is detected, the design is classified as combinational. Otherwise, STG performs FSM detection to distinguish FSM-dominated designs from general sequential circuits.
FSM detection uses two complementary methods:
\textbf{(1) Deterministic pattern matching.} STG scans the HDL source for \texttt{always\_ff} (or \texttt{always @(posedge clk)}) blocks that contain \texttt{case}/\texttt{casez} statements indexed by a register whose name matches common state-variable patterns (e.g., \texttt{state}). If such a pattern is found, the design is classified as FSM-dominated.
\textbf{(2) LLM-assisted analysis.} When deterministic matching is inconclusive, STG issues a structured prompt to an LLM, providing the module source and requesting a JSON description of the FSM, including state variables, encodings, static parameters, and transitions and the associated conditions. This step extracts the FSM structure needed for targeted stimulus generation (\S\ref{subsec:fsm}).
If neither method identifies an FSM, the design is classified as general sequential.

\textbf{\textit{Signal classification.}}
\begin{table}[t]
  \centering
  \caption{Signal classification heuristics. All roles use LCS-based fuzzy matching against name hints (case-insensitive).}
  \label{tab:signal-class}
  \vspace{-1em}
  \footnotesize
  \begin{tabular}{llr}
    \toprule
    \textbf{Role} & \textbf{Name hints} & \textbf{Width} \\
    \midrule
    Clock   & \texttt{clk}, \texttt{clock}                              & 1\,b \\
    Reset   & \texttt{rst}, \texttt{reset}, \texttt{rst\_n}             & 1\,b \\
    \midrule
    Control & \texttt{en}, \texttt{sel}, \texttt{start}, \texttt{load}, \texttt{valid}, \texttt{op} ...& $\leq$\,8\,b \\
    Data    & \texttt{d}, \texttt{data}, \texttt{din}, \texttt{addr} ...    & $\geq$\,8\,b \\
    \bottomrule
  \end{tabular}
  \vspace{-2em}
\end{table}

After determining the design type, STG classifies each input port into one of four roles: \emph{clock}, \emph{reset}, \emph{control}, or \emph{data} (output ports are handled uniformly by the checking logic). Table~\ref{tab:signal-class} summarizes the heuristics. All four categories use longest common subsequence (LCS)-based fuzzy matching: each port name is compared against category-specific hint lists using a LCS similarity score. Clock and reset signals are identified first and the remaining input signals are classified as control or data using the same LCS-based matching against their respective hint lists (Table~\ref{tab:signal-class}), combined with a width-based heuristic: narrow signals receive a higher control score, while wide signals receive a higher data score. When scores are tied, the signal defaults to control. In interactive mode, users may override any classification by providing explicit signal-role mappings.

\subsection{Testbench Generation Strategies}
\label{subsec:stim}

All three strategies share a common template-based architecture: STG renders testbenches from parameterized \textit{Jinja} templates, filling in module names, port lists, signal roles, and strategy-specific parameters. Based on the classified design type, STG selects the appropriate stimulus strategy---exhaustive-control enumeration for combinational designs, two-pass clocked stimulus for general sequential designs, or FSM traversal for FSM-dominated designs---and populates the corresponding template. Each generated testbench instantiates both the DUT and the golden reference with shared input drivers and separate output wires, and invokes a unified comparison task after every stimulus event. STG also handles per-output error counters, which are accumulated throughout the simulation and reported as pass rates at the end.

\subsubsection{Combinational}
\label{subsec:comb}

For combinational designs, the testbench applies stimulus directly without a clock. Following the signal partition in Table~\ref{tab:signal-class}, control signals are enumerated exhaustively over all $2^{b_c}$ combinations (where $b_c$ is the total control-input width), and for each control vector, data signals are randomized independently over $N_s$ samples. The total number of test vectors is therefore $2^{b_c} \times N_s$. After each stimulus application, the testbench invokes the comparison task to check all outputs. To keep simulation cost bounded, STG enforces a configurable upper bound $2^{b_{\max}}$ on the total vector count, requiring $2^{b_c} \times N_s \le 2^{b_{\max}}$; $N_s$ is automatically reduced when this bound would otherwise be exceeded. This strategy ensures that every reachable control mode is tested, while data-path behavior within each mode is sampled with high probability. For designs whose total input width is small enough, treating all inputs as control signals effectively yields exhaustive verification.

\subsubsection{General Sequential}
\label{subsec:seq}

Sequential designs require clock-driven simulation and careful handling of resets and output timing. A key issue is that a sequential design may use synchronous or asynchronous resets, and its outputs may follow Moore semantics (changing only on clock edges) or Mealy semantics (changing combinationally in response to inputs within a cycle). A testbench that only checks outputs at the positive clock edge may miss Mealy-style output changes, while one that ignores asynchronous reset behavior may miss recovery bugs.
This issue is not well handled by AutoBench and its follow-ups~\cite{Liu2024AutoBench,Liu2025CorrectBench,Liu2025ConfiBench}, where the LLM is asked to generate clock-based input stimulus and a Python-based checker that consumes the DUT outputs cycle by cycle. That structure naturally assumes observation only at clock boundaries: the generated checker receives one output snapshot per cycle, rather than intermediate within-cycle responses. As a result, Mealy-style behaviors that depend on intra-cycle input changes are easily overlooked even if the clocked trace appears correct.

\begin{figure}[t]
  \centering
  \includegraphics[width=0.95\columnwidth]{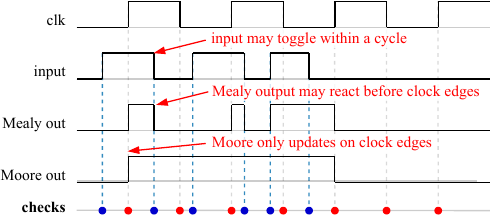}
  \vspace{-1em}
  \caption{Timing structure of the general sequential strategy. The signal labeled ``mealy'' corresponds to a latch-like within-cycle response, while the signal labeled ``moore'' corresponds to an FF-based registered response. STG inserts comparison points after intra-cycle input changes as well as at negative and positive edges, so both behaviors are observed.}
  \label{fig:seq-timing}
  \vspace{-1.5em}
\end{figure}

As illustrated in Fig.~\ref{fig:seq-timing}, STG addresses these issues through a two-pass, reset-aware strategy with multi-phase comparison points. Inputs are driven inside the clock period rather than only at the period boundary, and outputs are checked after intra-cycle input changes and at the clock edges. This allows the testbench to observe both within-cycle reactions and edge-triggered updates, so Mealy-style behavior is not missed while Moore-style registered behavior is still verified. The same framework also handles reset recovery: in the first pass, no resets are injected, allowing the design to accumulate state under sustained stimulus, whereas in the second pass resets are probabilistically inserted between stimulus cycles. The reset task adapts to the reset type, asserting and releasing the signal at clock boundaries for synchronous resets and exercising short assert--deassert sequences for asynchronous resets.

The stimulus generation in the sequential strategy follows a two-level randomization structure rather than a single ``enumerate control, then randomize data'' loop. Following Table~\ref{tab:signal-class}, STG first performs outer-loop random data injection before control enumeration, allowing the design to accumulate state under unconstrained data activity and exposing behaviors that are sensitive to prior history. It then enumerates control inputs exhaustively over all $2^{b_c}$ combinations. For each control vector, STG performs an inner loop of data randomization, repeatedly sampling data inputs while holding the control setting fixed. As a result, the stimulus schedule can be viewed as \emph{random data} $\rightarrow$ \emph{control assignment} $\rightarrow$ \emph{random data}, rather than a single flat sampling loop.

\subsubsection{FSM-Guided}
\label{subsec:fsm}

For FSM-dominated designs, random stimulus is unlikely to reach deep states or exercise rare transitions within a practical number of cycles. STG uses the extracted FSM structure (\S\ref{subsec:module-parsing}) to guide stimulus generation toward full transition coverage.

\begin{figure}[t]
  \centering
  \includegraphics[width=0.95\columnwidth]{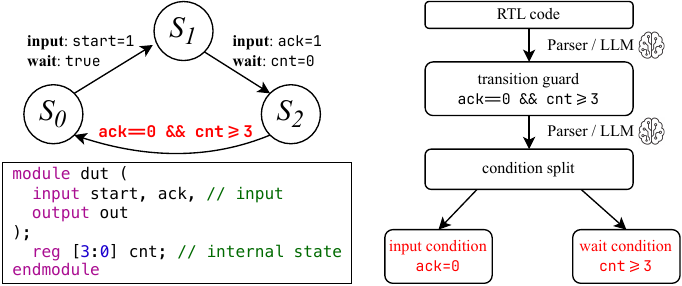}
  \vspace{-1em}
  \caption{Example of FSM-guided traversal. STG separates directly drivable input signals from internal wait conditions.}
  \label{fig:fsm-traversal}
  \vspace{-1.5em}
\end{figure}

As shown in Fig.~\ref{fig:fsm-traversal}, the FSM-guided strategy operates in two stages. In the \emph{generation stage}, STG extracts a state-transition graph from the DUT via deterministic pattern matching or LLM-assisted analysis (\S\ref{subsec:module-parsing}), and generates a C++ testbench that encodes the graph. Each edge guard is split into an \emph{input condition} (predicates over drivable ports) and a \emph{wait condition} (predicates over internal runtime state). For example, \texttt{ack==0\ \&\&\ cnt>=3} becomes: drive \texttt{ack=0}, and wait until \texttt{cnt>=3}. This separation allows STG to drive controllable inputs deterministically while internal conditions are satisfied naturally. The C++ testbench is then compiled together with the Verilog DUT and golden reference into a single executable via Verilator, which translates Verilog modules into C++ classes and thereby enables high-level constructs such as recursive traversal.

In the \emph{simulation stage}, the harness traverses the graph by DFS. At each state, STG parses the input condition into a lightweight AST and performs deterministic constraint extraction to derive concrete signal assignments (e.g., resolving \texttt{ack==0} to \texttt{ack=0}). It then drives those assignments and advances the clock until the wait condition is satisfied or a timeout is reached. If a transition is infeasible or times out, STG resets and backtracks to explore an alternative path, systematically covering all reachable transitions without random exploration. To verify that each transition is genuinely exercised at the HDL level, we extend Verilator's coverage API to expose per-line execution counts at runtime, providing a fine-grained signal for whether the HDL statements associated with the target transition have actually been reached. Finally, the testbench reports both pass rates and transition-coverage statistics at the end.

\subsection{Extension to the Silver-Reference Setting}
This paper focuses on the known-reference setting, where a trusted golden implementation is available. Nevertheless, STG's structural-analysis and stimulus-generation pipeline is not inherently tied to this assumption: the golden HDL module can be replaced by a software reference model (the \emph{silver reference} mentioned in Section~\ref{subsec:llm-tb}), typically emitted as a C++ or SystemC header.

\begin{figure}[t]
  \centering
  \includegraphics[width=0.9\columnwidth]{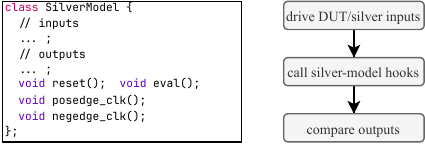}
  \vspace{-1em}
  \caption{Simplified structure of the silver-reference template. STG generates a C++ interface with DUT-aligned inputs and outputs, while the LLM fills in the behavioral logic.}
  \label{fig:silver-template}
  \vspace{-2em}
\end{figure}

Fig.~\ref{fig:silver-template} illustrates this extension. STG generates a skeleton software-model interface whose fields mirror the DUT ports and whose hooks align with the testbench's event structure. The LLM only needs to fill in the behavioral logic inside this fixed interface; STG preserves the same stimulus schedule and comparison flow used in the golden-reference setting. The C++ code is then compiled with Verilog files via Verilator, which enables the conversion of Verilog modules into C++ classes. Because verification quality now depends on the LLM-generated reference rather than a trusted golden implementation, this mode trades oracle reliability for broader applicability. We include it here to show that STG's architecture generalizes beyond the known-reference setting evaluated in this work.

  \section{Applications of STG}
\label{sec:applications}

The STG framework described in Section~\ref{sec:methodology} is not limited to a single benchmark format. More generally, it provides a structured verification backend for LLM-driven RTL workflows whenever the main bottleneck is reliable stimulus generation and low-cost behavioral checking. We highlight three representative applications.

\textbf{\textit{Replacing ad hoc benchmark testbenches.}}
Benchmark suites such as VerilogEval~\cite{Liu2023VerilogEval,Thakur2024RevisitingVerilogEval} and CVDP~\cite{pinckney2025cvdp} typically rely on hand-written verification artifacts. STG can be used directly by benchmark designers as a testbench-construction interface: given the DUT and reference, it generates a working testbench shell with the appropriate structure for combinational, sequential, or FSM-dominated designs. This is useful even in interactive mode, where a human can provide signal-role hints or design-type overrides and then build on top of the generated scaffold.

In practice, this means benchmark authors do not need to write every testbench from scratch. STG can quickly provide the module instantiation, clock/reset handling, and default stimulus structure, after which a human can add extra corner-case patterns or benchmark-specific checks if needed. This reduces manual effort while keeping the final benchmark testbench extensible rather than fully opaque or LLM-generated end-to-end.

\textbf{\textit{Verification-oriented data curation.}}
As outlined in Section~\ref{subsec:data-curation}, model-distillation pipelines generate large numbers of candidate DUTs that must be filtered before they can serve as training data~\cite{Yao2025CodeV,QiMeng2025CodeVR1,teng2025verirl,Chen2026SiliconMindV1}. Filtering is still commonly handled through prompt-based LLMs or LLM-generated verification artifacts, which are expensive and difficult to scale for large datasets such as PyraNet~\cite{nadimi2025pyranet}.

\begin{figure}[t]
  \centering
  \includegraphics[width=\columnwidth]{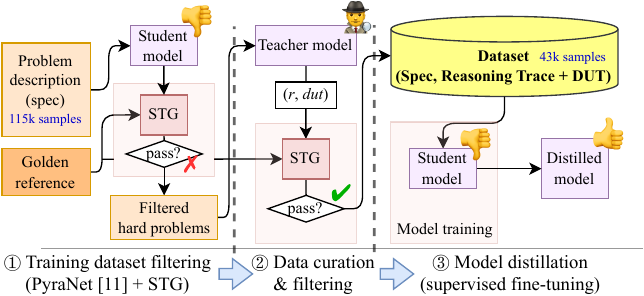}
  \vspace{-2.1em}
  \caption{Verification-oriented data-curation and training flow. (1) filters the source dataset to retain hard problems; (2) generates candidate DUTs with a teacher model and verifies with STG; (3) trains the student model on the curated data.}
  \label{fig:training-flow}
  \vspace{-2.1em}
\end{figure}

Fig.~\ref{fig:training-flow} shows our simple three-step data-curation and SFT workflow with STG. We start from a pool of 692k PyraNet samples and first down-select about 115k candidates using problem-difficulty and code-quality indicators provided by the source dataset due to limited computational resources. In Step~(1), STG is used to identify hard problems that are not already solved by the small base models, and samples correctly solved by the base models are removed. In Step~(2), a teacher model generates a reasoning trace and Verilog answer for each remaining problem, and STG again uses the golden reference to verify whether the teacher-produced DUT is correct. After this verification-based curation stage, 43k samples remain. In Step~(3), the surviving samples are used to train the student model.

STG plays two distinct roles in this workflow. First, it acts as a difficulty filter by measuring which problems remain unsolved by small base models, allowing us to focus the curation budget on informative training targets. Second, it serves as the verifier for teacher-generated answers, retaining only correct solutions. This makes STG a practical screening engine for large-scale RTL data curation before SFT, bypassing the need for per-problem LLM invocation required by recent specialized RTL models~\cite{Chen2026SiliconMindV1,teng2025verirl,QiMeng2025CodeVR1}.

\textbf{\textit{Verification backend for test-time scaling.}}
As discussed in Section~\ref{subsec:data-curation}, recent RTL generation systems increasingly use iterative search and refinement at inference time~\cite{wei2026vflow,min2026revolution,Dong2025ScaleRTL}. In these systems, verification is no longer a one-shot final check but part of the optimization loop: the quality of each search iteration depends directly on the quality of the verification signal.

\begin{figure}[t]
  \centering
  \includegraphics[width=\columnwidth]{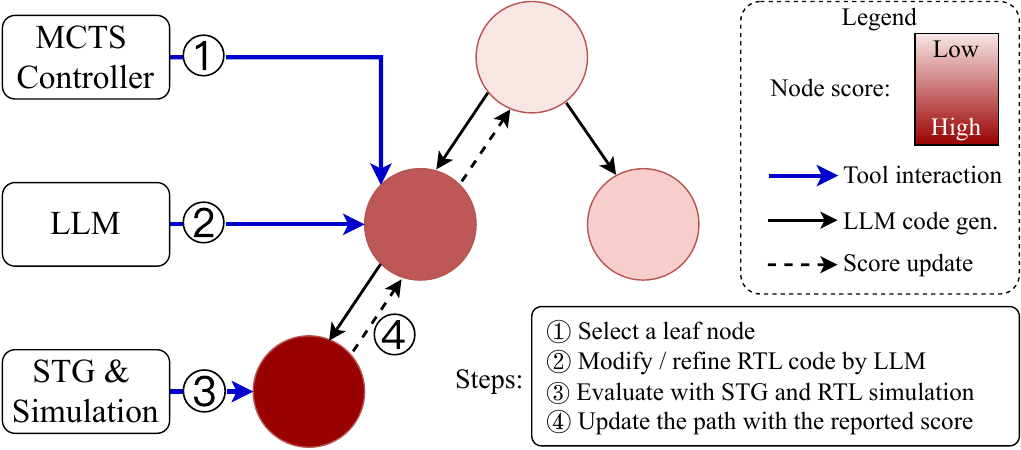}
  \vspace{-2.3em}
  \caption{Modified MCTS-based refinement flow with STG as the verification backend.}
  \label{fig:mcts-flow}
  \vspace{-1em}
\end{figure}

Fig.~\ref{fig:mcts-flow} shows our modified MCTS-style refinement loop based on VFlow~\cite{wei2026vflow}. Starting from a selected leaf node, the LLM proposes a modified RTL candidate, which is then verified by STG through testbench generation and RTL simulation. The reported score is propagated back along the search path and used to guide subsequent node selection. In this flow, STG serves as a drop-in replacement for the benchmark-provided testbench.
Compared with a fixed benchmark testbench, STG explores a wider set of scenarios and provides more concrete feedback about candidate behavior. This gives the search loop a stronger signal, allowing it to reject weak candidates earlier, guide refinement more effectively, and reach correct designs with fewer iterations and lower token cost.

  \begin{figure}[t]
  \centering
  \includegraphics[width=0.9\columnwidth]{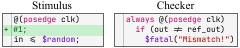}
  \vspace{-1.3em}
  \caption{Race condition in VerilogEval testbenches and its fix.
    We manually insert \texttt{\textbf{\color{teal}\#1}} (highlighted) after the clock edge.}
  \label{fig:verilogeval-fix}
  \vspace{-1.5em}
\end{figure}

\section{Experimental Results and Evaluation}
\label{sec:experiments}
In this section, STG is evaluated across the three application scenarios described in Section~\ref{sec:applications}: \textbf{(1) Testbench quality and DUT classification (\S\ref{subsec:exp-tb-quality})}: STG is benchmarked against a ConfiBench-style~\cite{Liu2025ConfiBench} iterative LLM testbench generation pipeline on VerilogEval, followed by a coverage analysis contrasting STG's sequential-random and FSM-guided strategies on a deep-state FSM design.
\textbf{(2) Verification-oriented data curation (\S\ref{subsec:exp-curation})}: STG serves as the verification engine for large-scale training-data filtering, and the resulting distilled models are evaluated against state-of-the-art specialized small language models.
\textbf{(3) Test-time scaling (\S\ref{subsec:exp-evolution})}: STG replaces the benchmark-provided testbench as the verification backend in an MCTS-based code refinement loop, and STG's search efficiency is measured across four backbone language models.

\vspace{-0.5em}
\subsection{Experimental Setup}
\label{subsec:setup}

All LLM inference experiments use GPT-OSS-120B~\cite{openai2025gptoss} running on one NVIDIA GB200 GPU. STG's deterministic pipeline (parsing, signal classification and template rendering) runs on a single CPU core (Intel Xeon w9-3475X, max 4.8\,GHz); for FSM-dominated designs, STG additionally invokes GPT-OSS-120B to extract the state-transition graph. For model distillation, training data is sourced from PyraNet~\cite{nadimi2025pyranet}, a large-scale dataset for RTL generation training; we use GPT-OSS-120B as the teacher and fine-tune three student models---Qwen2.5-Coder-7B-Instruct, Qwen3-4B-Thinking, and Qwen3-8B~\cite{hui2024qwen25codertechnicalreport,yang2025qwen3technicalreport}---on 16 NVIDIA H100 GPUs. Our training recipe is intentionally simple: after STG-based data curation, each student is trained with only an SFT stage. We compare against recent specialized small LMs that use more complex fine-tuning pipelines, including multi-stage SFT (SiliconMind-V1~\cite{Chen2026SiliconMindV1}) and combined SFT and RL methods (CodeV-R1~\cite{QiMeng2025CodeVR1} and VeriRL~\cite{teng2025verirl}). Importantly, this comparison is not driven by giving STG newer training data than previous works. For test-time scaling, we use four backbone LLMs spanning a wide range of model sizes and training recipes: SiliconMind-V1-7B~\cite{Chen2026SiliconMindV1}, GPT-OSS-120B~\cite{openai2025gptoss}, DeepSeek-R1-FP4-685B~\cite{Guo2025deepseek}, and one of our STG-curated distilled models.

We use Verilator~\cite{Snyder2024Verilator}, an open-source Verilog simulator, to perform RTL simulations; line and toggle coverage metrics are collected through Verilator's built-in coverage instrumentation.
All three experiment tracks use VerilogEval~\cite{Liu2023VerilogEval,Thakur2024RevisitingVerilogEval} (156 problems). The model-distillation experiments (\S\ref{subsec:exp-curation}) additionally evaluate on RTLLM-v2~\cite{lu2024rtllm} (50 problems) and CVDP~\cite{pinckney2025cvdp} categories cid02 and cid03 (172 problems), which cover non-agentic code completion and generation tasks suited to our target: RTL generation. CVDP is a newer and harder benchmark that is not used by prior works~\cite{QiMeng2025CodeVR1,teng2025verirl}.

\textbf{\textit{Note on VerilogEval testbench correctness.}}
During our evaluation, we identified cases where both manual inspection and STG agreed that a generated DUT was functionally correct, yet VerilogEval's original testbench reported a failure. The root cause is a race condition: as shown in Fig.~\ref{fig:verilogeval-fix}, both the stimulus and checker blocks trigger on the same clock edge with no ordering guarantee, so the checker may compare a newly driven input against a stale reference-model output. The fix is to insert a single \texttt{\#1} delay in the stimulus block after the clock edge, ensuring all reference evaluations complete before new inputs are driven. All VerilogEval results reported in this paper use our manually corrected testbenches.

\begin{table}[t]
  \centering
  \begin{minipage}[t]{0.50\columnwidth}
    \centering
    \captionsetup{type=table}
    \caption{Testbench generation comparison on VerilogEval.}
    \vspace{-1.3em}
    \label{tab:tb-gen-comparison}
    \resizebox{0.9\columnwidth}{!}{
    \setlength{\tabcolsep}{2.5pt}
    \begin{tabular}{@{}lccc@{}}
      \toprule
      \textbf{Method} & \textbf{Time} & \textbf{Line} & \textbf{Toggle} \\
      \midrule
      Pure-LLM & 92.4\,s & 93.97\% & 85.40\% \\
      \textbf{STG (Ours)} & \textbf{0.13\,s} & \textbf{95.88\%} & \textbf{95.77\%} \\
      \bottomrule
    \end{tabular}
    }
  \end{minipage}
  \hfill
  \begin{minipage}[t]{0.45\columnwidth}
    \centering
    \captionsetup{type=table}
    \caption{DUT classification accuracy.}
    \vspace{-1.3em}
    \label{tab:classification}
    \setlength{\tabcolsep}{3pt}
    \resizebox{\columnwidth}{!}{
    \begin{tabular}{@{}lrc@{}}
      \toprule
      \textbf{Outcome} & \textbf{Num.} & \textbf{Percentage} \\
      \midrule
      Both correct & 2{,}734 & 89.8\% \\
      LLM \ding{51}, STG \ding{55} & 27 & \ 0.9\% \\
      LLM \ding{55}, STG \ding{51} & 236 & \ 7.8\% \\
      Both wrong & 49 & \ 1.6\% \\
      \bottomrule
    \end{tabular}
    }
  \end{minipage}
  \vspace{-1.5em}
\end{table}

\begin{table*}[t]
  \centering
  \caption{Pass@k (\%) before and after training, grouped by base model. We report pass@k with $n \text{ (number of samples)}=20$.}
  \label{tab:training}
  \vspace{-1em}
  \resizebox{0.99\textwidth}{!}{
  \begin{tabular}{l|lcc||ccc|ccc|ccc||c}
    \toprule
    \multirow{2}{*}{\textit{\textbf{Role}}}& \multicolumn{1}{c}{\multirow{2}{*}{\textbf{Model}}} & \multirow{2}{*}{\textbf{SFT}} & \multicolumn{1}{c||}{\multirow{2}{*}{\textbf{RL}}} & \multicolumn{3}{c|}{\textbf{RTLLM-v2~\cite{lu2024rtllm}}} & \multicolumn{3}{c|}{\textbf{VerilogEval-v2~\cite{Liu2023VerilogEval,Thakur2024RevisitingVerilogEval}}} & \multicolumn{3}{c||}{\textbf{CVDP~\cite{pinckney2025cvdp}}} & \textbf{Z-score (\%)} \\
    & & & & \textbf{p@1} & \textbf{p@5} & \textbf{p@10} & \textbf{p@1} & \textbf{p@5} & \textbf{p@10} & \textbf{p@1} & \textbf{p@5} & \textbf{p@10} & \textbf{p@10} \\
    \midrule
    \textit{Teacher} & GPT-OSS-120B~\cite{openai2025gptoss} & -- & -- & 69.9 & 78.1 & 80.8 & 89.6 &  96.7 & 97.6 & 42.9 & 57.9 & 62.2 & 94 \\
    \midrule
    \multirow{3}{*}{\textit{Base}}
      & Qwen2.5-C-7B-Instruct~\cite{hui2024qwen25codertechnicalreport}  & -- & -- & 29.3 & 48.6 & 56.0 & 33.6 & 53.7 & 60.1 & 13.6 & 25.1 & 29.8 & -151 \\
      & Qwen3-4B-Thinking~\cite{yang2025qwen3technicalreport}      & -- & -- & 36.4 & 50.9 & 56.3 & 21.4 & 30.4 & 33.4 & 15.4 & 24.8 & 29.1 & -200 \\
      & Qwen3-8B~\cite{yang2025qwen3technicalreport}               & -- & -- & 40.2 & 61.1 & 67.6 & 52.5 & 65.4 & 69.1 & 17.4 & 28.7 & 34.4 & -80 \\
    \midrule
    \multirow{12}{*}{\textit{Fine-tuned}}
      & \multicolumn{3}{l||}{\textit{Base: Qwen2.5-C-7B-Instruct}} \\
      & \ \ CodeV-R1~\cite{QiMeng2025CodeVR1}            & \checkmark & \checkmark & \cellcolor{bronze!50}68.0 & \cellcolor{bronze!50}\textbf{77.6} & \cellcolor{silver!50}\textbf{80.7} & 73.2 & 83.6 & 86.6 & \cellcolor{bronze!50}\textbf{34.5} & 50.4 & 54.8 & 54 \\
      & \ \ VeriRL (paper)~\cite{teng2025verirl}                & \checkmark & \checkmark & 63.3 & 70.3 & -- & 67.2 & 76.1 & -- & -- & -- & -- & -- \\
      & \ \ \quad$\hookrightarrow$\, \, (reproduced) & \checkmark & \checkmark & \cellcolor{gold!50}\textbf{71.8} & \cellcolor{bronze!50}\textbf{77.6} & 78.8 & 60.8 & 74.8 & 78.8 & 18.1 & 27.5 & 31.9 & -29 \\
      & \ \ SiliconMind-V1~\cite{Chen2026SiliconMindV1}      & \checkmark & $\times$ & 63.8 & 74.0 & 75.9 & \textbf{73.9} & 83.6 & 85.8 & 31.3 & 47.5 & 52.9 & 30 \\
      & \ \ \textbf{STG (Ours)}                     & \checkmark & $\times$ & 63.1 & 76.4 & 79.0 & 70.5 & \textbf{84.9} & \textbf{89.4} & 32.4 & \cellcolor{bronze!50}\textbf{50.5} & \cellcolor{bronze!50}\textbf{56.0} & \cellcolor{bronze!50}\textbf{56} \\
    \addlinespace
      & \multicolumn{3}{l||}{\textit{Base: Qwen3-4B-Thinking}} \\
      & \ \ SiliconMind-V1~\cite{Chen2026SiliconMindV1}      & \checkmark & $\times$ & \textbf{67.9} & 75.3 & 76.0 & \cellcolor{gold!50}\textbf{82.0} & 89.6 & 91.0 & 33.4 & 47.3 & 51.9 & 37 \\
      & \ \ \textbf{STG (Ours)}                    & \checkmark & $\times$ & 67.5 & \cellcolor{silver!50}\textbf{78.2} & \cellcolor{bronze!50}\textbf{79.8} & 80.0 & \cellcolor{gold!50}\textbf{90.2} & \cellcolor{bronze!50}\textbf{91.5} & \cellcolor{silver!50}\textbf{35.6} & \cellcolor{silver!50}\textbf{52.4} & \cellcolor{silver!50}\textbf{57.9} & \cellcolor{silver!50}\textbf{68} \\
    \addlinespace
      & \multicolumn{3}{l||}{\textit{Base: Qwen3-8B}} \\
      & \ \ SiliconMind-V1~\cite{Chen2026SiliconMindV1}      & \checkmark & $\times$ & 66.6 & 74.9 & 76.5 & \cellcolor{silver!50}\textbf{81.0} & \cellcolor{bronze!50}89.8 & \cellcolor{gold!50}\textbf{92.4} & 34.4 & 49.2 & 53.8 & 46 \\
      & \ \ \textbf{STG (Ours)}                    & \checkmark & $\times$ & \cellcolor{silver!50}\textbf{68.7} & \cellcolor{gold!50}\textbf{79.9} & \cellcolor{gold!50}\textbf{81.9} & \cellcolor{bronze!50}80.2 & \cellcolor{silver!50}\textbf{89.9} & \cellcolor{silver!50}92.0 & \cellcolor{gold!50}\textbf{36.5} & \cellcolor{gold!50}\textbf{52.9} & \cellcolor{gold!50}\textbf{58.1} & \cellcolor{gold!50}\textbf{77} \\
    \bottomrule
  \end{tabular}
  }
  \parbox{\textwidth}{\footnotesize\raggedright
  \textbf{The final column reports the mean Z-score of pass@10 across the three benchmarks as a single aggregate metric: for each benchmark we compute $z = (x - \mu)/\sigma$, then average the three resulting Z-scores.}
  Colors denote rankings among all fine-tuned models: \colorbox{gold!50}{first}, \colorbox{silver!50}{second}, and \colorbox{bronze!50}{third}. Bold marks the best within each base-model group.}
  \vspace{-1.5em}
\end{table*}

\vspace{-0.5em}
\subsection{Testbench Quality and DUT Classification}
\label{subsec:exp-tb-quality}

We first evaluate STG as a direct replacement for human-crafted testbenches on VerilogEval. For each of the 156 problems, we use GPT-OSS-120B to generate approximately 10 correct and 10 incorrect variants from the golden reference, yielding 3{,}046 DUTs in total. Each DUT is verified by two methods: (1)~\emph{Pure-LLM}, a ConfiBench-style~\cite{Liu2025ConfiBench} prompt-based testbench with up to 5 iterative refinement rounds, and (2)~\emph{STG}, a single-pass STG-generated testbench.

Table~\ref{tab:tb-gen-comparison} summarizes the generation cost and coverage metrics. STG generates testbenches $\mathbf{720\times}$ faster than the iterative LLM approach while achieving higher line and toggle coverage (+1.9 and +10.4 pp): STG exhaustively enumerates all combinations of control-flow signals, guaranteeing that every control path is exercised at least once for combinational designs, whereas the stochastic LLM testbench may leave rare control states untested.

Table~\ref{tab:classification} breaks down the classification outcomes into four categories. STG and the LLM-based testbench agree on 91.4\% of cases. In the 7.8\% of cases where only STG succeeds, the dominant failure mode is the LLM testbench producing a false PASS on an incorrect DUT (193 out of 236 cases), confirming that stochastic testbenches are unreliable at detecting subtle bugs. The remaining failures, 0.9\% where only STG fails and 1.6\% where both fail, share the same reason: bugs that require exhaustive state-space enumeration to expose, beyond the reach of either structured or stochastic stimulus.

\begin{figure}[t]
  \centering
  \includegraphics[width=0.8\columnwidth]{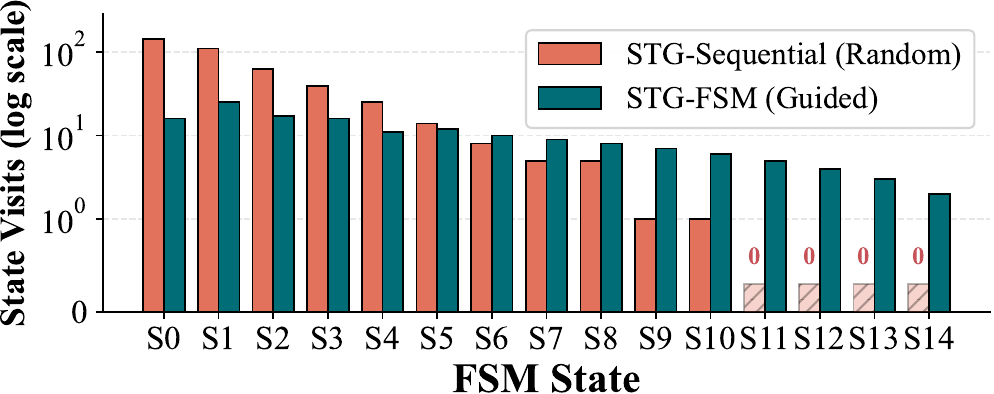}
  \vspace{-1.2em}
  \caption{State visit counts under STG-Sequential (random stimulus) and STG-FSM (guided traversal) for a 15-state Mealy sequence detector. Random stimulus visits decay exponentially and fail to reach states S11--S14.}
  \label{fig:coverage-comparison}
\vspace{-1em}
\end{figure}

\subsubsection{Coverage on FSM-Dominated Designs}
\label{subsec:exp-fsm-coverage}

To illustrate when the FSM-guided strategy (\S\ref{subsec:fsm}) is most valuable, we compare the two STG modes on a 15-bit sliding-window Mealy sequence detector with 15 states (S0--S14) and 30 transitions. This design requires a specific 15-bit input sequence to trigger the detection output, a scenario where random stimulus is exponentially unlikely to succeed.
Fig.~\ref{fig:coverage-comparison} shows the per-state visit counts. Under STG-Sequential with random stimulus, visits decay exponentially and states S11--S14 are never entered. In contrast, STG-FSM performs DFS passes over the extracted transition graph, achieving 100\% transition coverage. This result shows that FSM-guided traversal is essential for designs with deep state spaces that random stimulus cannot penetrate. In the main experiments in Table~\ref{tab:tb-gen-comparison}, all 156 VerilogEval problems are verified using the general sequential strategy, which already achieves high coverage on the benchmark's predominantly shallow-state designs. The FSM-guided mode serves as a complementary strategy for a subset of designs where targeted state exploration is required.

\begin{table}[t]
  \centering
  \caption{Resource comparison for testbench generation on 115k problems: pure-LLM (GB200) vs.\ STG (a CPU core).}
  \vspace{-1em}
  \small
  \label{tab:curation-scaling}
  \begin{tabular}{lrr}
    \toprule
    \textbf{Metric} & \textbf{Pure-LLM} & \textbf{STG} \\
    \midrule
    Runtime                & 59.1\,h   & 5.6\,h    \\
    Hardware power         & 1{,}200\,W & ${\approx}$\,100\,W  \\
    Total energy           & 70.9\,kWh & 0.56\,kWh \\
    Hardware cost          & \$60{,}000 & \$4{,}000  \\
    Compilation rate       & 71.3\%    & 100\%     \\
    \bottomrule
  \end{tabular}
  \vspace{-1.3em}
\end{table}

\begin{figure*}[t]
  \begin{minipage}[c]{0.18\textwidth}
    \centering
    \vspace{2em}
    \resizebox{\textwidth}{!}{%
    \begin{tabular}{lcc}
      \toprule
      \textbf{Model} & \textbf{w/o} & \textbf{w/} \\
      \midrule
      STG-Qwen3-4B      & 94.9 & \textbf{96.2} \\
      SiliconMind-V1-7B  & 91.7 & \textbf{92.3} \\
      GPT-OSS-120B       & \textbf{97.4} & \textbf{97.4} \\
      DeepSeek-R1-685B   & \textbf{98.1} & \textbf{98.1} \\
      \bottomrule
    \end{tabular}%
    }
    \captionof{table}{Pass rate (\%) at 256 search nodes.}
    \label{tab:pass-rate-256}
  \end{minipage}%
  \hfill
  \begin{minipage}[c]{0.80\textwidth}
    \centering
    \includegraphics[width=\linewidth]{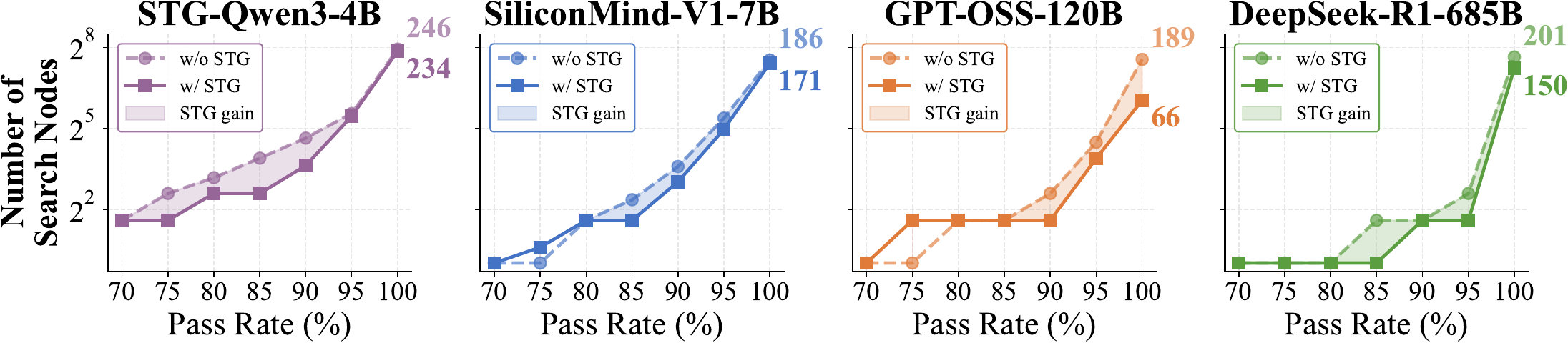}
    \vspace{-2.5em}
    \captionof{figure}{Percentage of correctly solved problems vs.\ search node budget for four backbone models.}
    \label{fig:evolution-scaling}
  \end{minipage}
  \vspace{-1.3em}
\end{figure*}

\begin{figure}[t]
  \centering
  \includegraphics[width=\columnwidth]{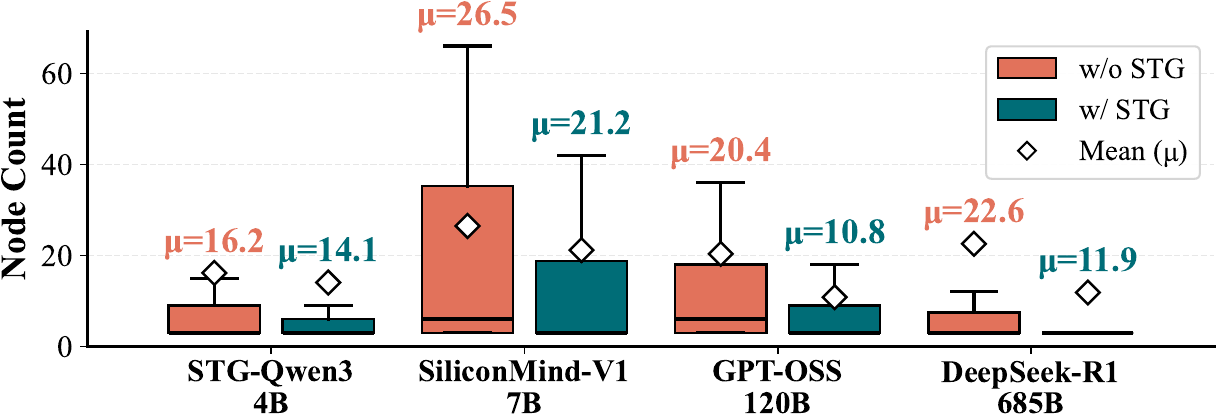}
  \vspace{-2em}
  \caption{Node-count distribution for non-trivial and solved problems for each model. Outliers beyond $1.5\times$ the interquartile range (IQR) are suppressed for readability.}
  \label{fig:ablation-efficiency}
  \vspace{-1.5em}
\end{figure}

\vspace{-0.5em}
\subsection{Verification-Oriented Data Curation}
\label{subsec:exp-curation}

We evaluate STG as the verification engine for large-scale data curation in a model-distillation pipeline, as illustrated in Fig.~\ref{fig:training-flow}.

Table~\ref{tab:curation-scaling} compares the resource footprint of testbench generation on 115k problems. The pure-LLM baseline uses single-pass generation on an GB200 GPU without iterative refinement, of which only 71.3\% produce compilable testbenches, while STG guarantees compilable output by construction. STG on a single CPU core completes the task in 5.6 hours compared to 59.1 hours for the LLM baseline ($10.6\times$ speedup). Because STG runs on a CPU core (${\approx}$100\,W) rather than a 1{,}200\,W GPU, STG provides total energy reduction by  $127\times$ (from $70.9$ to $0.56$ kWh), on hardware that costs 15$\times$ less. Moreover, STG's pipeline is trivially parallelizable via CPU multiprocessing for further speedup with minimal engineering effort.

\textit{\textbf{Model training results.}}
Table~\ref{tab:training} presents our fine-tuning results, grouped by base model to facilitate direct comparison. We report $\text{pass}@k = \mathbb{E}\!\left[1 - \binom{n-c}{k}\!/\binom{n}{k}\right]$, the unbiased estimator of the probability that at least one of $k$ samples passes, where $n$ is the total number of generated samples and $c$ is the number of successful ones. As a single aggregate metric across the three benchmarks, our STG-trained models achieve the top three mean Z-scores for pass@10. Despite relying on only a single SFT stage after STG-based data curation, our models remain competitive with or outperform more complex multi-stage SFT~\cite{Chen2026SiliconMindV1} and SFT+RL~\cite{QiMeng2025CodeVR1,teng2025verirl} pipelines. On Qwen2.5-Coder-7B-Instruct, our model surpasses previous work on VerilogEval and CVDP at pass@5 and pass@10. On the Qwen3 series, our models achieve the strongest CVDP results and the best pass@5/pass@10 on VerilogEval within each base-model group. While RL-based methods perform well on RTLLM (a 2024 benchmark), their complexity is not justified by consistent gains on the newer 2025 benchmarks, VerilogEval-v2 and CVDP.

We also encountered substantial reproducibility issues with VeriRL. Relative to the numbers presented in the paper~\cite{teng2025verirl}, our replicated VeriRL checkpoint scores significantly higher on RTLLM-v2 but worse on VerilogEval-v2 even after applying our VerilogEval testbench fix, whereas the other evaluated models consistently improve under the corrected benchmark. Combined with VeriRL's weak transfer to VerilogEval-v2 and CVDP, this discrepancy suggests that the released model may overfit artifacts specific to RTLLM rather than delivering robust gains across newer benchmarks.

Overall, the results demonstrate that a simple data curation pipeline powered by STG can yield strong and competitive distilled models with only one simple SFT stage, without the need for complex multi-stage SFT and RL-centric training workflows.

\vspace{-0.5em}
\subsection{Test-Time Scaling}
\label{subsec:exp-evolution}

We integrate STG into an MCTS-based test-time scaling refinement loop based on VFlow~\cite{wei2026vflow}, as illustrated in Fig.~\ref{fig:mcts-flow}, and compare it against using the benchmark-provided testbench as the verification oracle. Experiments are conducted on our modified VerilogEval with four backbone LLMs: three prior models (SiliconMind-V1-7B, GPT-OSS-120B, DeepSeek-R1-685B) and our STG-curated distilled model from Section~\ref{subsec:exp-curation} (STG-Qwen3-4B-Thinking). For each problem, the search expands nodes until the candidate DUT passes the testbench or a budget of 256 nodes is exhausted.

Table~\ref{tab:pass-rate-256} reports the pass rate at the full 256-node budget, and Fig.~\ref{fig:evolution-scaling} shows the number of search nodes required to reach each pass-rate percentile in the 70--100\% range. Across all four backbone LLMs, STG matches or improves the pass rate (Table~\ref{tab:pass-rate-256}) and reduces the node count at most percentiles (Fig.~\ref{fig:evolution-scaling}). Fig.~\ref{fig:ablation-efficiency} further details the node-count distribution for non-trivial solved problems (i.e., those requiring more than one search node), showing that STG lowers the mean node count by 14--47\% and compresses both the interquartile range and median count. Because STG tests more patterns and reports per-output-port pass rates, the verification signal is more informative and guides LLM to search more efficiently. Overall, STG's contribution to test-time scaling is twofold: it increases the final pass rate and reduces per-problem search cost.
\vspace{-0.5em}
  \section{Conclusion}
\label{sec:conc}

This paper presents STG, a structured testbench generation framework that treats module-level RTL verification as a structured generation problem rather than unconstrained code synthesis.
Powered by design type-specific template-based rendering, STG produces testbenches deterministically at $720\times$ the speed of iterative LLM approaches with higher coverage.
Across three application scenarios, STG consistently outperforms LLM-based alternatives at a fraction of the cost: it detects 7.8\% more incorrect DUTs, reduces MCTS search node count by 14--47\% on large backbone models, and enables large-scale data curation $11\times$ faster on a single CPU core than LLM-based filtering while supporting strong distilled models with only a SFT stage.
These results establish STG as a practical, low-cost verification backbone for LLM-driven HDL workflows, also suggesting that the effectiveness of recent complex RL training workflows remains questionable, especially on newer benchmarks where our simpler pipeline provides competitive performance.

Future work includes integration with reliable FSM extraction for complex production RTL. Additionally, as LLM-driven hardware design moves toward continuous learning---where models are iteratively retrained on newly generated data---efficient and reliable data curation becomes increasingly critical; STG's low-cost verification pipeline is well positioned to support such end-to-end workflows. Finally, the strong HDL-specialized small language models produced by STG-curated distillation are natural candidates for speculative decoding, where a lightweight draft model accelerates inference of a larger backbone while preserving exact output quality.

  \vspace{-0.5em}
\begin{acks}
We acknowledge the financial support from Academia Sinica's SiliconMind Project (AS-IAIA-114-M11). This work was also supported in part by the National Science and Technology Council, Taiwan (112-2221-E-002-159-MY3), as well as the National Center for High-performance Computing and Taipei-1 for computational resources.
\end{acks}
  \newpage
  \bibliographystyle{ACM-Reference-Format}
  \bibliography{refs.bib}

  \end{document}